\documentclass{article}

\usepackage{microtype}
\usepackage{graphicx}
\usepackage{subcaption}
\usepackage{booktabs} 
\usepackage{graphicx}  
\usepackage{booktabs}  
\usepackage{multirow}  
\usepackage{makecell}  
\usepackage{bm} 
\usepackage{enumitem} 

\usepackage{hyperref}

\usepackage[accepted]{icml2025}

\usepackage{amsmath}
\usepackage{amssymb}
\usepackage{mathtools}
\usepackage{amsthm}

\usepackage[capitalize,noabbrev]{cleveref}

\theoremstyle{plain}

\theoremstyle{definition}

\theoremstyle{remark}

\usepackage[textsize=tiny]{todonotes}

\def\MLP{\mathrm{MLP}}
\def\EMLP{\mathrm{EquivMLP}}

\icmltitlerunning{Torsional-GFN}

\begin{document}

\twocolumn[
\icmltitle{Torsional-GFN: a conditional conformation generator for small molecules}



\icmlsetsymbol{equal}{*}

\begin{icmlauthorlist}
\icmlauthor{Léna Néhale Ezzine}{equal,udem,mila}
\icmlauthor{Alexandra Volokhova}{equal,udem,mila}
\icmlauthor{Piotr Gaiński}{ju}
\icmlauthor{Luca Scimeca}{udem,mila}
\icmlauthor{Emmanuel Bengio}{rec}
\icmlauthor{Prudencio Tossou}{rec}
\icmlauthor{Yoshua Bengio}{udem,mila,cifar}
\icmlauthor{Alex Hernandez-Garcia}{udem,mila}
\end{icmlauthorlist}

\icmlaffiliation{mila}{Mila -- Quebec AI Institute}
\icmlaffiliation{udem}{Université de Montréal}
\icmlaffiliation{ju}{Jagiellonian University}
\icmlaffiliation{rec}{Valence Labs, Recursion}
\icmlaffiliation{cifar}{CIFAR}

\icmlcorrespondingauthor{Alexandra Volokhova}{alexandra.volokhova@mila.quebec}
\icmlcorrespondingauthor{Léna Néhale Ezzine}{lena-nehale.ezzine@mila.quebec}

\icmlkeywords{Machine Learning, ICML}

\vskip 0.3in
]



\printAffiliationsAndNotice{\icmlEqualContribution} 

\begin{abstract}
Generating  stable molecular conformations is crucial in several drug discovery applications, such as estimating the binding affinity of a molecule to a target. 
Recently, generative machine learning methods have emerged as a promising, more efficient method than molecular dynamics for sampling of conformations from the Boltzmann distribution.
In this paper, we introduce Torsional-GFN, a conditional GFlowNet specifically designed to sample conformations of molecules proportionally to their Boltzmann distribution, using only a reward function as training signal. Conditioned on a molecular graph and its local structure (bond lengths and angles), Torsional-GFN samples rotations of its torsion angles.
Our results demonstrate that Torsional-GFN is able to sample conformations approximately proportional to the Boltzmann distribution for multiple molecules with a single model, and allows for zero-shot generalization to unseen bond lengths and angles coming from the MD simulations for such molecules.
Our work presents a promising avenue for scaling the proposed approach to larger molecular systems, achieving zero-shot generalization to unseen molecules, and including the generation of the local structure into the GFlowNet model.

\end{abstract}

\section{Introduction}

The problem of sampling stable molecular configurations corresponds to generating independent molecular conformations $c \in C_G$ from the Boltzmann distribution $p(c) \propto \exp(-E(c)/k_b T)$  with known but unnormalized density.
For thermodynamic calculations, a more complete sampling of the conformations provides a better estimate of important quantities in drug discovery, such as free energy, and binding affinity to a target \cite{molani2024accurate}. 

Among computational chemistry methods, molecular dynamics is the \textit{de facto} approach, where methods such as CREST have shown promise in their ability to accurately capture low-energy conformations \cite{pracht2020automated}. However, despite the accuracy of these methods, it remains computationally expensive for high throughput applications and large compounds. Although there is ongoing research in faster alternatives with knowledge-based algorithms, such as distance-geometry approaches like ETKDG \cite{riniker2015better}, these methods cannot sample in accordance to the Boltzmann distribution.

Recently, generative models have emerged as a promising approach for conditional sampling of conformations from the Boltzmann distribution.
For instance, Boltzmann generators \cite{noe2019boltzmann} train normalizing flows which map from easy-to-sample distributions, like standard Gaussians, to the energy landscape of interest. Training typically consists of both data- and energy-based training. Data-based training  involves maximizing the likelihood of the observed data (most often using the forward-KL loss), while energy-based training minimizes the reverse-KL loss between the flow's learned density and the unnormalized target Boltzmann density;\ however, reverse-KL and forward-KL formulations typically suffer from mode-seeking and mean-seeking behaviors, respectively, which can adversely impacting training dynamics and sample quality \cite{malkin2022gflownets}. Although learning with off-policy distributions is possible with re-weighted importance sampling, it often introduces a high gradient variance.
Recently, GFlowNets \cite{bengio2023gflownet} have emerged as a new class of generative models that can be trained to sample according to a reward function. Further, they can be trained using any behavior policy without introducing high gradient variance \cite{malkin2022gflownets}, making them well-suited for the problem of exploring the energy landscape of molecules.

In this paper, we introduce Torsional-GFN, a method  to sample conformations of a molecule conditioned on the molecular graph and the local structure (i.e.\ bond lengths and angles). Torsional-GFN is a conditional GFlowNet that generates conformations of a given molecular graph by sampling rotations of its torsion angles.
We summarize the contributions of our work as follows:
\begin{itemize}[nosep]
    \item We demonstrate the ability of a single GFlowNet model to approximate the Boltzmann distribution of multiple molecules, building on previous work \cite{volokhova2024towards} which required training one GFlowNet per molecular system, 
    \item we introduce a new graph neural network architecture, VectorGNN, designed to infer geometrical information about the torsion angles in a molecule's 3D structure,
    \item we demonstrate the ability of GFlowNets, trained on a dataset of six molecules with fixed bond length and angle, to generalize to unseen values of the bond lengths and angles from MD simulations of new molecules.
    
\end{itemize}

\section{Method}

\subsection{Preliminaries}

Adopting the definitions and notations from \citet{jing2022torsional}, we consider a molecular graph $G = (V, E)$, where nodes $V$ are the atoms and edges $E$ are the atom bonds. $C_G$ denotes the space of possible conformations.
 
A \textbf{conformation} $c \in C_G$ is a set of $\mathrm{SE}(3)$-equivalent vectors in $\mathbb{R}^{3|V|}$. In other words, $C_G$ maps to the quotient space $\mathbb{R}^{3|V|} / \mathrm{SE}(3)$. In this space, two vectors that are equal up to an $\mathrm{SE}(3)$-transformation belong to the same equivalence class, and are therefore considered to be the same conformation.

Our goal is to draw independent conformations $c \in C_G$ from the Boltzmann distribution:\  
\begin{equation}\label{eq:boltzmann}
    p(c | G) = \frac{1}{Z(G)}\exp\left(\frac{-E(c)}{k_bT}\right),
\end{equation}
where $E(c)$ is the internal energy of conformation $c$, $Z(G)$ is a normalization constant for the molecular graph $G$, $k_B$ is the Boltzmann constant, and $T$ is the absolute temperature.

The definitions above describe the space of conformations in terms of an extrinsic (or Cartesian) coordinate frame. A conformation $c \in C_G$ can also be specified in terms of its \textit{intrinsic coordinates}: 
\begin{itemize}[nosep]
    \item Bond lengths and bond angles, referred to as the molecule's local structure are denoted by $L$,
    \item Torsion angles, consisting of dihedral angles around freely rotatable bonds, are denoted by $\Phi$ (see \cref{tasparam}).
\end{itemize}
 
Previous works \cite{volokhova2024towards, jing2022torsional} either rely on the assumption that the local structure of stable conformations is constant \cite{hawkins2017conformation, riniker2015better}, or they sample local structures from \citet{rdkit} cheminformatics software . However, we argue that, if we care about the full potential energy landscape, such approximations do not hold, as shown in \cref{md_rdkit_localstructures} on a dataset of 8 molecules. In particular, RDKit significantly underestimates the bond lengths vibrations ( \cref{fig:bonglenthshist}).

\subsection{Torsional-GFN}


Given a molecular graph $G$, we construct 3D conformations $c \in C_G$ by sampling both the local structure $L \sim p(L|G)$ and the rotatable torsion angles $\Phi \sim p(\Phi | L,G)$. This allows us to write $c = c(L, \Phi)$, making the intrinsic coordinates explicit, and to decompose the Boltzmann distribution from \cref{eq:boltzmann} into
\begin{equation}\label{eq:pext2int}
    p(c | G) =   \underbrace{\frac{1}{\sqrt{\mathrm{det}(g)}} p(\Phi | G, L)}_{\text{learned by Torsional-GFN}} \underbrace{p (L | G)}_{\text{\quad sampled from MD}},   
\end{equation}
where $\frac{1}{\sqrt{\mathrm{det}(g)}}$ describes the volume change due to the conversion from intrinsic to extrinsic coordinates. We set  $\mathrm{det}(g) = 1$ for simplicity and offer a detailed discussion in \cref{likelihoodintr2extr}. 

To reduce the dimensionality of our search space, following \citet{volokhova2024towards}, we let the GFlowNet learn the distribution of torsional angles $p(\Phi | G, L)$ and obtain the local structures $p (L | G)$ from MD simulations. Note that $p(\Phi | G, L) = \frac{1}{Z(G,L)} \exp\left(- \frac{E(c(\Phi, L))}{k_BT}\right)$. As such, we can define the  \textit{unnormalized} reward function for the vector of torsion angles $\Phi$ as:\ 
\begin{equation}\label{eq:reward}
    R(\Phi| G, L) = \exp\left(- \frac{E(c(\Phi, L))}{k_BT}\right).
\end{equation}

We train a GFlowNet model, which is designed to learn sampling probabilities $p_\top^{\theta} (\Phi | G, L)$ proportional to the reward in \cref{eq:reward}.

Torsional-GFN is a continuous GFlowNet \citep{lahlou2023theory}, conditioned on molecules defined by their molecular graph and local structure $(G, L)$. Given a molecule with $m$ rotatable torsion angles, $\Phi =  (\phi^1, \dots, \phi^m)$, Torsional-GFN samples these torsion angles in the hypertorus $\mathcal{X} = [0, 2\pi]^m$ \citep{volokhova2024towards}. The sampling process starts from a source state $\Phi_0$ and continues with a trajectory of sequential updates $\tau = (\Phi_0 \rightarrow \Phi_1 \rightarrow \dots \rightarrow \Phi_n = \Phi)$ according to a trainable forward policy $P_F(\Phi_{t} | \Phi_{t - 1})$. Torsional-GFN also uses a trainable backward policy $P_B(\Phi_{t - 1} | \Phi_t)$. Similarly as in \citet{volokhova2024towards}, we parametrize the forward and backward policies of Torsional-GFN as a mixture of von Mises distributions with learnable mixing weights $w$, locations $\mu$ and concentrations $\kappa$ (details in \cref{supp:vonmises}). Given a batch of trajectories $B_{\tau}$ using any behavior policy, we train the GFlowNet to minimize the Vargrad loss \citep{richter2020vargrad}: 
\begin{align*}\label{eq:vargrad}
    & \mathcal{L}_{\mathrm{VG}}(B_{\tau};\theta | G, L)= \mathbb{E}_{\tau \in B_{\tau}} [ \log(Z_\theta(G, L))  \\
    & -  \log \underbrace{\frac{P_B^\theta( \tau  \mid \Phi_n, G, L) R(\Phi_n | G, L)}{P_F^\theta(\tau | G, L)} }_{\mathcal{C}_{\theta}(\tau | G, L)}  ]^2,
\end{align*}
where $\log Z_{\theta}(G, L) =  \mathbb{E}_{\tau' \in B_{\tau}}   
 \log \mathcal{C}_\theta(\tau' | G, L)$. Thus, we can train Torsional-GFN on a dataset $D$ of molecular graphs with their local structures using the following loss: 
\begin{equation}
    \mathcal{L}_D = \mathbb{E}_{G_i, L_i \sim D} \mathcal{L}_{\mathrm{VG}}(B_{\tau};\theta | G_i, L_i).
\end{equation}

Note that the behavior policy, used to sample trajectories and take gradient steps on $\mathcal{L}_D$ can be \emph{any full-support off-policy distribution} (see \cref{alg:torsionalGFNalgo}), unlike in Boltzmann generators \citep{noe2019boltzmann}, where the behavior policy is restricted to either the forward or reverse-KL loss. Moreover, Torsional-GFN does not require any importance sampling \citep{jing2022torsional}, thus avoiding the issues of bias and high-variance that come with it.

\subsection{VectorGNN}
In order to train a single conditional GFlowNet with multiple molecular systems---unlike previous work \citep{volokhova2024towards}, where a separate GFlowNet was trained for each molecular system with a policy parametrized with an MLP---we introduce a new graph-neural network architecture, VectorGNN, for the Torsional-GFN policy models. We found it to be faster than other GNNs such as the one used in \citet{jing2022torsional}.

A mixture of von Mises distributions with components $k \in 1, \ldots K$ is defined with locations $\mu_k(\phi)$, concentrations $\kappa_k(\phi)$ and mixture weights $w_k(\phi)$. The location parameter determines the expected direction of torsion angle updates, and must thus be \textit{reflection-equivariant}. VectorGNN learns forces that generate a torque for each rotatable bond, then outputs a set of pseudo-scalars $o_\phi \in \mathbb{R}^{K \times 3}$, which are are mapped to the distribution parameters as follows (further details in \cref{vectorgnn}):\
\[
w_k(\phi) = (o_\phi)^2_{k,1}, \quad
\mu_k(\phi) = (o_\phi)_{k,2}, \quad
\kappa_k(\phi) = (o_\phi)^2_{k,3}.
\]

\section{Experiments}
\subsection{Setup}

We train Torsional-GFN on a subset of six molecules, each with $m = 2$ rotatable torsion angles, from the FreeSolv dataset \cite{mobley2014freesolv}. To investigate generalization to unseen molecules, we use two other molecules as a test set . The first transition $\Phi_0 \rightarrow \Phi_1$ is sampled from the uniform distribution on $[0,2\pi]^m$ and the total length of the trajectory is $n = 6$.

For evaluation, we run MD simulations to obtain a dataset of 2001 ground truth conformations from the Boltzmann distribution for each molecule. 
During training, the bond lengths and angles for each molecule were fixed to the values of one arbitrary conformation from the MD simulation dataset. 
We provide more detail in Appendix \ref{seq:app_md}.

As energy function, we use the MMFF94s \citep{halgren1999mmff}, a widely-used force field that provides a good estimation of the internal energy for small organic molecules.

In order to accelerate the training process, we pre-train VectorGNN on the supervised task of predicting the energy and torsion angles values of molecular conformations, as detailed in \cref{supp:pretraining}.   
Then, we use the pretrained model as an initialization for training both $P_F^{\theta}$ and $P_B^{\theta}$.
\subsection{Results}

Our main objective for the evaluation is to investigate how closely the torsion angles from Torsional-GFN align with those from the true Boltzmann distribution, and whether the model can generalize to unseen local structures and unseen molecules. 

\paragraph{Proximity to the reward landscape}
To compare the sampling distribution of the Torsional-GFN $p_\top^{\theta}(\Phi |G, L)$ with the reward $R(\Phi | G, L)$, we discretize the 2D torus to a grid of 10,000 uniformly sampled points $\{\Phi_i\}_{i=1}^{10000} \in [0,2\pi]^2$. Then, we estimate $\log p_\top^{\theta}(\Phi_i | G, L)$ and compute $\log R(\Phi_i | G, L)$ on the grid. 
This way, we can compute the correlation coefficient between the corresponding logarithmic values $\rho_{\log p_\top^{\theta}, log R}$ using \cref{eq:corr_coeff}, see \cref{seq:app_corr} for more details.


In addition, we estimate the discretized target distribution $ P(\Phi_i | G, L) =  R(\Phi_i | L, G) / \sum_k  R(\Phi_k | L, G)$, and the discretized sampling distribution $ P_\top^{\theta}(\Phi_i | G, L) = p_\top^{\theta}(\Phi_i | L, G) / \sum_k p_\top^{\theta}(\Phi_k | L, G)$, and compare them using the Jenson-Shannon divergence $\mathrm{ JSD}\Big(P_\top^{\theta}(\Phi_i | G, L) ||  P(\Phi_i | G, L)\Big)$, that we denote as $\mathrm{JSD}^P$ in Table \ref{tab:metrics}.

\begin{figure}[h]
\centering
\begin{subfigure}{0.23\textwidth}
  \includegraphics[width=\textwidth]{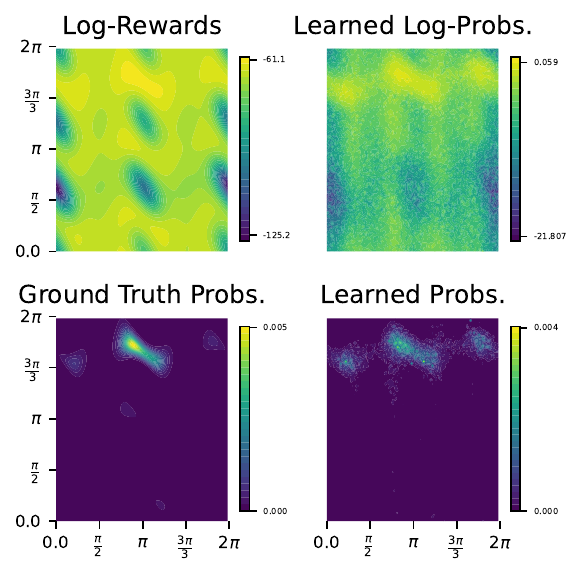}
  \caption{Train local structure}
  \label{fig:2d_train}
\end{subfigure}
\begin{subfigure}{0.23\textwidth}
  \includegraphics[width=\textwidth]{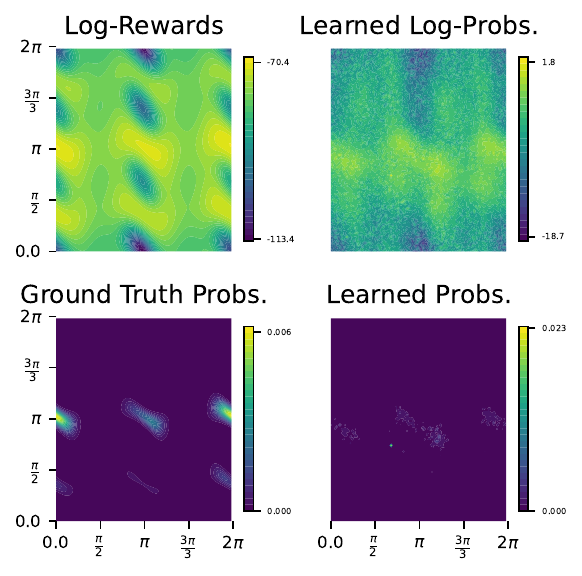}
  \caption{Unseen local structure}
  \label{fig:2d_unseen}
\end{subfigure}
\caption{Visualization of the log-rewards and ground truth probabilities alongside the learned sampling distributions for molecule \texttt{COc1ccccc1}.}
\label{fig:2d-main}
\end{figure}

\cref{fig:2d-main} shows the log-reward landscape and the estimated log-probabilities for one of the train molecules, as well as the ground truth probabilities and estimated probabilities, for the train local structure in \cref{fig:2d_train} and for an unseen local structure in \cref{fig:2d_unseen}. For this molecule, the metrics are close to the median value over the training set of molecules (see \cref{fig:2dviztrain_mol1,fig:2dviztrain_mol2,fig:2dviztrain_mol3,fig:2dviztrain_mol4,fig:2dviztrain_mol5} for visualizations of the results for other molecules. \cref{tab:metrics} provides the values of $\mathrm{JSD}^p$ and  $\rho_{\log p_\top^{\theta}, log R}$ for the train and unseen local structures of the 6 train and 2 test molecules. 

Both visualizations and quantitative results suggest that Torsional-GFN is able to learn the overall reward landscape for most of the train molecules and train local structures. We also observe that our model captures the main features of the landscape for the unseen local structures. By comparing \cref{fig:2d_unseen} and \cref{fig:2d_train}, we clearly see that Torsional-GFN is able to follow the shift of the modes in the energy landscape when sampling torsion angles for unseen local structures. Finally, in the results for the test molecules with corresponding unseen local structures (see \cref{fig:2dviztest} and the last 2 rows of \cref{tab:metrics}), while we observe that for one molecule our model sampling distribution is very far from the target, for the other test molecule we notice limited coverage of some modes and low probability regions of the energy landscape. This does not demonstrate consistent generalization to the test molecules, but indicates a potential for it.

\begin{table}[htbp]
\centering
\scriptsize
\resizebox{\columnwidth}{!}{%
\begin{tabular}{lccccc}
\hline
\textbf{SMILES} & $\mathrm{JSD}^E_{\mathrm{GFN}} / \mathrm{JSD}^E_{rand}$ & \multicolumn{2}{c}{$\mathrm{JSD}^{p}$} & \multicolumn{2}{c}{$\rho_{\log p_T, \log R}$} \\
& & \textbf{Train} & \textbf{Unseen} & \textbf{Train} & \textbf{Unseen} \\
\hline
C1C=CC[C@@H]2[C@@H]1... & 0.3613 & 0.4097 & 0.6845 & 0.5493 & 0.3011 \\
COC=O & 0.0201 & 0.0358 & 0.0911 & 0.7068 & 0.7042 \\
C[C@@H]1CCCC[C@@H]1C & 0.1391 & 0.0270 & 0.2354 & 0.9561 & 0.6907 \\
COc1ccccc1 & 0.4255 & 0.1001 & 0.3484 & 0.6929 & 0.6338 \\
c1ccc2c(c1)C(=O)... & 0.0580 & 0.1417 & 0.2710 & 0.9269 & 0.9292 \\
CCC & 0.0433 & 0.0100 & 0.0256 & 0.9740 & 0.9242 \\
\hline
CCc1cccc2c1cccc2 & 0.9807 & N/A & 0.6206 & N/A & -0.0597 \\
c1ccc(c(c1)C(F)... & 0.3653 & N/A & 0.6795 & N/A & 0.3670 \\
\hline
\end{tabular}
}
\caption{Evaluation metrics. The top 6 rows correspond to train molecules and the bottom 2 rows are test molecules.}
\label{tab:metrics}
\end{table}

\begin{figure*}
    \centering
    \includegraphics[width=0.9\linewidth]{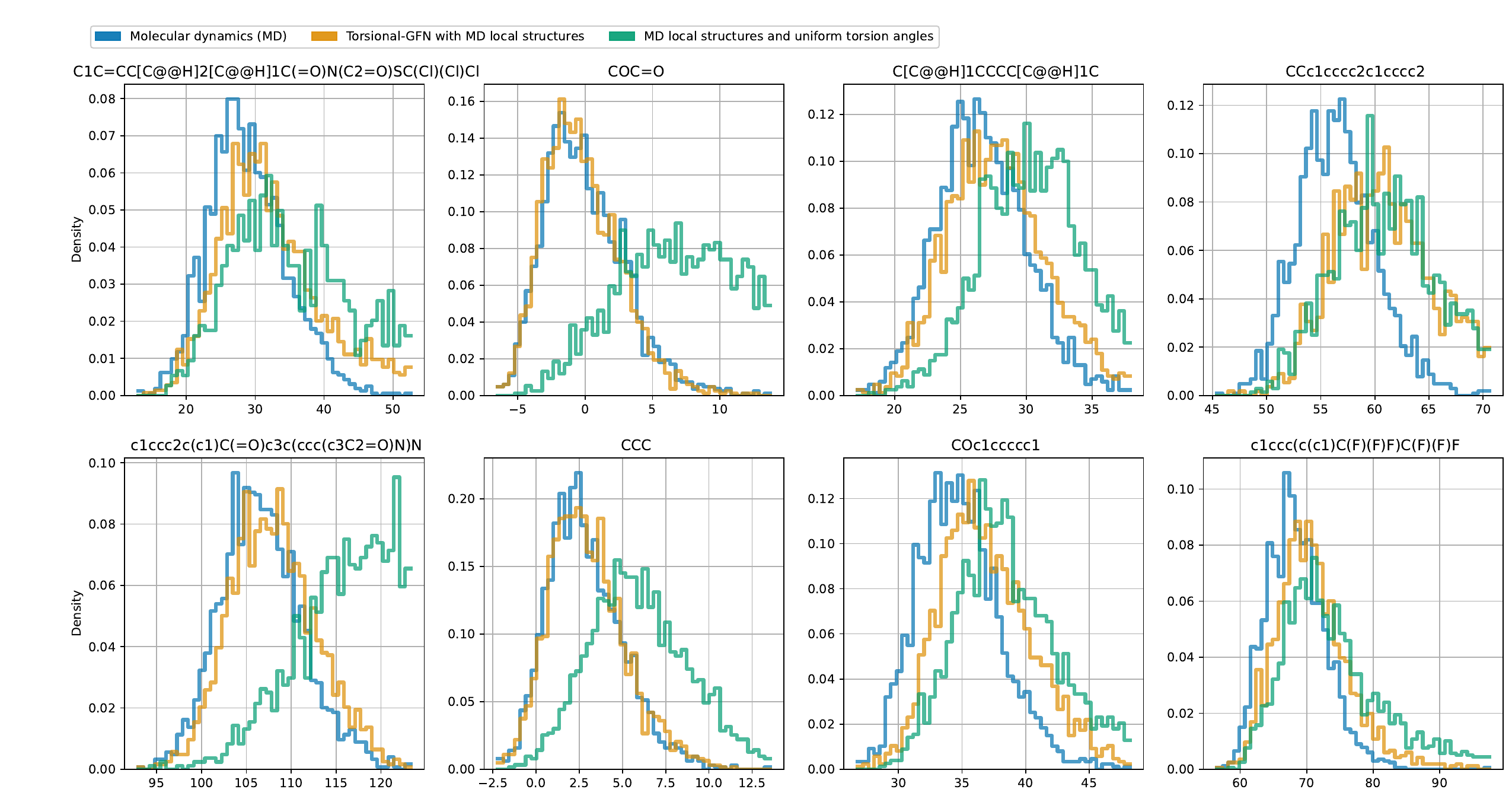}
    \caption{Energy histograms for the 6 train molecules (first three columns) and 2 test molecules (last column).}
    \label{fig:energy_hist}
\end{figure*}

\textbf{Energy histograms}
To further compare the sampling distribution with the target, we analyze the energy distribution of sampled conformations for each molecule. The target distribution here is given by the energies of the samples obtained with MD simulation. To obtain the corresponding distribution for Torsional-GFN, we sample one set of torsion angles for each local structure from MD simulation and then estimate the energy of the resulting conformations. 
For comparison, we also draw a random set of torsion angles from the uniform distribution on a torus. This results in three histograms per molecule, for which we restrict the energy range to the lower and upper bound of energies of MD.

To numerically measure the differences in energy distributions for each molecule, we normalize the energy histograms described above, resulting in three discrete probability distributions of the energies: $P_{MD}(E)$ for MD samples, $P_{GFN}(E)$ for Torsional-GFN samples, and $P_{rand}(E)$ for random samples. Then we compute Jenson-Shannon divergencies between MD energies and Torsional-GFN energies $\mathrm{JSD}^E_{\mathrm{GFN}} = \mathrm{JSD}\Big(P_{\mathrm{MD}}(E) || P_{\mathrm{GFN}}(E)\Big)$ and between MD energies and random energies $\mathrm{JSD}^E_{rand} = \mathrm{JSD}\Big(P_{\mathrm{MD}}(E) || P_{rand}(E)\Big)$. Finally, we divide the first by the latter and obtain the ratio  $\mathrm{JSD}^E_{\mathrm{GFN}}/\mathrm{JSD}^E_{rand}$. If this ratio is less than 1, this implies that Torsion-GFN distribution is closer to the target than random. The first column of table \cref{tab:metrics} shows the values of the ratio for each of the 6 train molecules and 2 test molecules. 

The values of the ratio metrics $\mathrm{JSD}^E_{\mathrm{GFN}}/\mathrm{JSD}^E_{rand}$ and the energy histograms in Figure \ref{fig:energy_hist} show that the energy distribution of Torsional-GFN samples is significantly closer to the MD energies than random for all the train molecules and one of the test molecules. 
Noticeably, for several train molecules, we observe that Torsional-GFN energies histogram overlap very closely with MD energies. 
This may suggest that the Torsional-GFN sampling distribution is close to the target distribution for many of the unseen local structures of the train molecules, which is consistent with the evaluation comparing probability landscapes described earlier in this section. 
Similarly, we observe that Torsional-GFN energies are very close to random for one of the test molecule, while for the other test molecule the model samples noticeably closer to the MD energies than random, pointing to potential generalization capacity to unseen molecules.



\section{Conclusion and future work}

In this work we proposed Torsional-GFN, a conditional GFlowNet method for sampling torsion angles of molecular conformations proportionally to the target distribution, and with conditioning on the input molecular graph and the local structure. Our experimental results suggest that Torsional-GFN is able to sample closely to the target distributions for the train molecules. However, some of them are more challenging to learn than others. Moreover, our method demonstrates generalization to unseen local structures and a potential for generalization to unseen molecules. 

We consider this work to be an intermediate milestone towards a GFlowNet-based sampler from the Boltzmann distribution which would be able to generalize to unseen molecules. In future work, we will explore possible ways of training Torsional-GFN on a bigger dataset of molecules with more torsion angles, sampling of the local structures with the GFlownet, and different ways to parametrize the policy distribution, e.g. with a simulation-free objective \cite{akhound2024iterated}.






\section*{Impact Statement}
The proposed method has a potential for speeding-up the sampling of small molecular conformations from the Boltzmann distribution, especially once current limitations are addressed by future work. This could affect various applications related to chemical properties prediction, such as drug discovery and materials design. We acknowledge that our work can also have potential malicious applications, such as chemical weapons development, and we firmly stand against these. Moreover, we acknowledge that if our work was used for potentially socially beneficial applications, such as drug discovery, it would be likely that only privileged big pharmaceutical companies would be able to use our method at scale and would capitalize on it, because training the Torsional-GFN is computationally expensive and requires a GPU with at least 40 GB of memory even for a small setting with 6 train molecules. Our deepest hope is that our work can contribute towards mitigating current and future pandemics, especially in underprivileged countries, which suffer the most from some classes of pandemics such as anti-microbial resistance. However, we acknowledge that such a contribution is very challenging in the context of unequal distribution of wealth in the world. 



\section*{Acknowledgments}
The authors thank Danyal Rehman for his contributions to our discussions and  Stephan Thaler for generating the MD dataset. We acknowledge Mila's IDT team support and Mila compute resources for necessary experiments.

L.S. acknowledges support from Recursion. Y.B. acknowledges support from CIFAR and the CIFAR AI Chair program as well as NSERC funding for the Herzberg Canada Gold medal. This project was undertaken thanks to funding from IVADO, the Canada First Research Excellence Fund and NRC AI4D.
\nocite{langley00}

\bibliography{cameraready_genbio25}

\begin{thebibliography}{24}
\providecommand{\natexlab}[1]{#1}
\providecommand{\url}[1]{\texttt{#1}}
\expandafter\ifx\csname urlstyle\endcsname\relax
  \providecommand{\doi}[1]{doi: #1}\else
  \providecommand{\doi}{doi: \begingroup \urlstyle{rm}\Url}\fi

\bibitem[Akhound-Sadegh et~al.(2024)Akhound-Sadegh, Rector-Brooks, Bose, Mittal, Lemos, Liu, Sendera, Ravanbakhsh, Gidel, Bengio, et~al.]{akhound2024iterated}
Akhound-Sadegh, T., Rector-Brooks, J., Bose, A.~J., Mittal, S., Lemos, P., Liu, C.-H., Sendera, M., Ravanbakhsh, S., Gidel, G., Bengio, Y., et~al.
\newblock Iterated denoising energy matching for sampling from boltzmann densities.
\newblock \emph{arXiv preprint arXiv:2402.06121}, 2024.

\bibitem[Axelrod \& Gomez-Bombarelli(2022)Axelrod and Gomez-Bombarelli]{axelrod2022geom}
Axelrod, S. and Gomez-Bombarelli, R.
\newblock Geom, energy-annotated molecular conformations for property prediction and molecular generation.
\newblock \emph{Scientific Data}, 9\penalty0 (1):\penalty0 185, 2022.

\bibitem[Bengio et~al.(2023)Bengio, Lahlou, Deleu, Hu, Tiwari, and Bengio]{bengio2023gflownet}
Bengio, Y., Lahlou, S., Deleu, T., Hu, E.~J., Tiwari, M., and Bengio, E.
\newblock Gflownet foundations.
\newblock \emph{The Journal of Machine Learning Research}, 24\penalty0 (1):\penalty0 10006--10060, 2023.

\bibitem[Boothroyd et~al.(2023)Boothroyd, Behara, Madin, Hahn, Jang, Gapsys, Wagner, Horton, Dotson, Thompson, et~al.]{boothroyd2023development}
Boothroyd, S., Behara, P.~K., Madin, O.~C., Hahn, D.~F., Jang, H., Gapsys, V., Wagner, J.~R., Horton, J.~T., Dotson, D.~L., Thompson, M.~W., et~al.
\newblock Development and benchmarking of open force field 2.0. 0: The sage small molecule force field.
\newblock \emph{Journal of Chemical Theory and Computation}, 19\penalty0 (11):\penalty0 3251--3275, 2023.

\bibitem[Eastman et~al.(2017)Eastman, Swails, Chodera, McGibbon, Zhao, Beauchamp, Wang, Simmonett, Harrigan, Stern, et~al.]{eastman2017openmm}
Eastman, P., Swails, J., Chodera, J.~D., McGibbon, R.~T., Zhao, Y., Beauchamp, K.~A., Wang, L.-P., Simmonett, A.~C., Harrigan, M.~P., Stern, C.~D., et~al.
\newblock Openmm 7: Rapid development of high performance algorithms for molecular dynamics.
\newblock \emph{PLoS computational biology}, 13\penalty0 (7):\penalty0 e1005659, 2017.

\bibitem[Halgren(1999)]{halgren1999mmff}
Halgren, T.~A.
\newblock Mmff vi. mmff94s option for energy minimization studies.
\newblock \emph{Journal of computational chemistry}, 20\penalty0 (7):\penalty0 720--729, 1999.

\bibitem[Hawkins(2017)]{hawkins2017conformation}
Hawkins, P.~C.
\newblock Conformation generation: the state of the art.
\newblock \emph{Journal of chemical information and modeling}, 57\penalty0 (8):\penalty0 1747--1756, 2017.

\bibitem[Hendrycks \& Gimpel(2016)Hendrycks and Gimpel]{hendrycks2016gaussian}
Hendrycks, D. and Gimpel, K.
\newblock Gaussian error linear units (gelus).
\newblock \emph{arXiv preprint arXiv:1606.08415}, 2016.

\bibitem[Jain et~al.(2022)Jain, Bengio, Hernandez-Garcia, Rector-Brooks, Dossou, Ekbote, Fu, Zhang, Kilgour, Zhang, et~al.]{jain2022biological}
Jain, M., Bengio, E., Hernandez-Garcia, A., Rector-Brooks, J., Dossou, B.~F., Ekbote, C.~A., Fu, J., Zhang, T., Kilgour, M., Zhang, D., et~al.
\newblock Biological sequence design with gflownets.
\newblock In \emph{International Conference on Machine Learning}, pp.\  9786--9801. PMLR, 2022.

\bibitem[Jing et~al.(2022)Jing, Corso, Chang, Barzilay, and Jaakkola]{jing2022torsional}
Jing, B., Corso, G., Chang, J., Barzilay, R., and Jaakkola, T.
\newblock Torsional diffusion for molecular conformer generation.
\newblock \emph{Advances in Neural Information Processing Systems}, 35:\penalty0 24240--24253, 2022.

\bibitem[Klein \& No{\'e}(2024)Klein and No{\'e}]{klein2024transferable}
Klein, L. and No{\'e}, F.
\newblock Transferable boltzmann generators.
\newblock \emph{arXiv preprint arXiv:2406.14426}, 2024.

\bibitem[Lahlou et~al.(2023)Lahlou, Deleu, Lemos, Zhang, Volokhova, Hern{\'a}ndez-Garc{\i}a, Ezzine, Bengio, and Malkin]{lahlou2023theory}
Lahlou, S., Deleu, T., Lemos, P., Zhang, D., Volokhova, A., Hern{\'a}ndez-Garc{\i}a, A., Ezzine, L.~N., Bengio, Y., and Malkin, N.
\newblock A theory of continuous generative flow networks.
\newblock In \emph{International Conference on Machine Learning}, pp.\  18269--18300. PMLR, 2023.

\bibitem[Langley(2000)]{langley00}
Langley, P.
\newblock Crafting papers on machine learning.
\newblock In Langley, P. (ed.), \emph{Proceedings of the 17th International Conference on Machine Learning (ICML 2000)}, pp.\  1207--1216, Stanford, CA, 2000. Morgan Kaufmann.

\bibitem[Malkin et~al.(2022)Malkin, Lahlou, Deleu, Ji, Hu, Everett, Zhang, and Bengio]{malkin2022gflownets}
Malkin, N., Lahlou, S., Deleu, T., Ji, X., Hu, E., Everett, K., Zhang, D., and Bengio, Y.
\newblock Gflownets and variational inference.
\newblock \emph{arXiv preprint arXiv:2210.00580}, 2022.

\bibitem[Mobley \& Guthrie(2014)Mobley and Guthrie]{mobley2014freesolv}
Mobley, D.~L. and Guthrie, J.~P.
\newblock Freesolv: a database of experimental and calculated hydration free energies, with input files.
\newblock \emph{Journal of computer-aided molecular design}, 28:\penalty0 711--720, 2014.

\bibitem[Molani \& Cho(2024)Molani and Cho]{molani2024accurate}
Molani, F. and Cho, A.~E.
\newblock Accurate protein-ligand binding free energy estimation using qm/mm on multi-conformers predicted from classical mining minima.
\newblock \emph{Communications Chemistry}, 7\penalty0 (1):\penalty0 247, 2024.

\bibitem[No{\'e} et~al.(2019)No{\'e}, Olsson, K{\"o}hler, and Wu]{noe2019boltzmann}
No{\'e}, F., Olsson, S., K{\"o}hler, J., and Wu, H.
\newblock Boltzmann generators: Sampling equilibrium states of many-body systems with deep learning.
\newblock \emph{Science}, 365\penalty0 (6457):\penalty0 eaaw1147, 2019.

\bibitem[Physics(2025)]{profoundphysics2025metric}
Physics, P.
\newblock The metric tensor: A complete guide with examples, 2025.
\newblock URL \url{https://profoundphysics.com/metric-tensor-a-complete-guide-with-examples/}.
\newblock Accessed: 2025-05-26.

\bibitem[Pracht et~al.(2020)Pracht, Bohle, and Grimme]{pracht2020automated}
Pracht, P., Bohle, F., and Grimme, S.
\newblock Automated exploration of the low-energy chemical space with fast quantum chemical methods.
\newblock \emph{Physical Chemistry Chemical Physics}, 22\penalty0 (14):\penalty0 7169--7192, 2020.

\bibitem[{RDKit}()]{rdkit}
{RDKit}.
\newblock Rdkit: Open-source cheminformatics.
\newblock \url{https://www.rdkit.org}.
\newblock Accessed: 2025-05-25.

\bibitem[Richter et~al.(2020)Richter, Boustati, N{\"u}sken, Ruiz, and Akyildiz]{richter2020vargrad}
Richter, L., Boustati, A., N{\"u}sken, N., Ruiz, F., and Akyildiz, O.~D.
\newblock Vargrad: a low-variance gradient estimator for variational inference.
\newblock \emph{Advances in Neural Information Processing Systems}, 33:\penalty0 13481--13492, 2020.

\bibitem[Riniker \& Landrum(2015)Riniker and Landrum]{riniker2015better}
Riniker, S. and Landrum, G.~A.
\newblock Better informed distance geometry: using what we know to improve conformation generation.
\newblock \emph{Journal of chemical information and modeling}, 55\penalty0 (12):\penalty0 2562--2574, 2015.

\bibitem[Vemgal et~al.(2023)Vemgal, Lau, and Precup]{vemgal2023empirical}
Vemgal, N., Lau, E., and Precup, D.
\newblock An empirical study of the effectiveness of using a replay buffer on mode discovery in gflownets.
\newblock \emph{arXiv preprint arXiv:2307.07674}, 2023.

\bibitem[Volokhova et~al.(2024)Volokhova, Koziarski, Hern{\'a}ndez-Garc{\'\i}a, Liu, Miret, Lemos, Thiede, Yan, Aspuru-Guzik, and Bengio]{volokhova2024towards}
Volokhova, A., Koziarski, M., Hern{\'a}ndez-Garc{\'\i}a, A., Liu, C.-H., Miret, S., Lemos, P., Thiede, L., Yan, Z., Aspuru-Guzik, A., and Bengio, Y.
\newblock Towards equilibrium molecular conformation generation with gflownets.
\newblock \emph{Digital Discovery}, 3\penalty0 (5):\penalty0 1038--1047, 2024.

\end{thebibliography}
\bibliographystyle{icml2025}

\newpage
\appendix
\onecolumn
\section{Appendix}



\subsection{Related work}

\paragraph{Molecular dynamics (MD) simulations} MD consists of running simulations of the evolution of the molecule's position over time using Newtonian physics. If we let MD run for a sufficient amount of time, then a Boltzmann-weighted ensemble of conformations can be obtained directly. This allows the calculation of any desired property of a molecule;\ however, to obtain uncorrelated samples from MD while maintaining stability, small update steps, on the order of femtoseconds, are typically needed. This makes MD simulations very slow, and more challenging when we have well-separated metastable states, where transitions are unlikely due to high-energy barriers \citep{klein2024transferable}.

\paragraph{Boltzmann generators}  In \citep{noe2019boltzmann}, the authors tackle the problem of sampling from an unnormalized Boltzmann density using normalizing flows. Training consists of both data-based and energy-based training, using respectively forward-KL loss and reverse-KL loss.  However, they suffer respectively from mode-seeking and mean-seeking behaviors, thus hurting training \cite{malkin2022gflownets}. Learning with off-policy distributions is possible with reweighted importance sampling, but it introduces a high gradient variance. The GFlowNet trajectory-based losses encompass these issues, and they can be trained using any behavior policy without introducing high gradient variance \cite{malkin2022gflownets}.

\paragraph{TorsionalDiffusion} In \citep{jing2022torsional}, which also serves as a basis of our work, a conditional diffusion model is trained to sample conformations of a molecule given its molecular graph as an input. To make the conformer search problem more manageable, authors apply a widely-used approximation for small molecules, called the rigid rotor (RR) approximation \citep{hawkins2017conformation}. In this framework, the bond lengths and bond angles are kept fixed, and the only degree of freedom is the torsion angles. TorsionalDiffusion is trained on QM9 and GEOM-DRUGS \citep{axelrod2022geom}, which provides standard conformer ensembles generated with metadynamics in CREST \citep{pracht2020automated}. Besides training with data, the authors propose an energy-based training approach to learn to sample from the corresponding Boltzmann distribution. However, this approach relies on reweighted importance sampling, yielding a variance that increases with the discrepancy between the learned policy and the true Boltzmann distribution 
 \cite{malkin2022gflownets}.

\paragraph{GFlowNets for molecular conformation generation} \citet{lahlou2023theory,volokhova2024towards} introduced GFlowNets for sampling conformations of small molecules from the Boltzmann distribution. In these earlier works, a dedicated GFlowNet with a MLP policy model was trained for each molecule to learn the potential energy surface. Here, we build upon these studies by introducing a conditional GFlowNet with a novel GNN architecture that serves as a policy model taking in the molecular graph as input. Moreover, we establish a framework for training a single GFlowNet on multiple molecules, potentially enabling amortized inference.

\subsection{Method details}

\subsubsection{Torsion angles parametrization}\label{tasparam}

As there are multiple ways to define torsion angles for a molecule \cite{jing2022torsional}, for each rotatable bond $(b_i, c_i)$, we choose arbitarily an atom $a_i$ in the neighborhood of $b_i$, and an atom $d_i$ in the neighborhood of $c_i$. Then, we define a torsion angle of bond $b_i, c_i$ using the quadruplet $a_i, b_i, c_i, d_i$. The arbitrary choice of $a_i, d_i$ does not affect our method: another choice $(a'_i, d'_i) \neq (a_i, d_i)$ would simply result in a translation (modulo $2 \pi$) of the reward landscape on the torus, making our method translation-equivariant to the choice of reference torsion angle. 

\subsubsection{Likelihood conversion from torsional space to $\mathbb{R}^{3|V|}/SE(3)$ }\label{likelihoodintr2extr}

Consider a molecule described by a vector $x \in \mathbb{R}^{3|V|}/SE(3)$, with $|V| \in \mathbb{N^+}$ the number of atoms. The Boltzmann distribution is usually computed in this space of extrinsic coordinates. However, we here operate on the space of \textit{intrinsic coordinates} $(L, \Phi)$. This change of variables implies the need to account for a factor change in the infinitesimal volume $dV$, when going from intrinsic to extrinsic space.  

As we are sampling the local structure $L$ directly from MD data, we can omit the volume change due to $L$, and restrict it to the space of rotatable torsion angles $\mathcal{X} = [0, 2 \pi]^m$. Note that if we were sampling $L$ with TorsionalGFN, we would need to account for the volume change due to $L$ as well.

The metric tensor $g$ is an object that describes the geometry of a coordinate system, that is its distances, angles and volumes \citep{profoundphysics2025metric}. Its components are dot products between basis vectors. For the molecule described above, the metric tensor $g(\mathbf{x})$ is defined as:
\begin{equation}
    g_{ij}(\mathbf{x}) = \langle\frac{d\mathbf{x}}{d\Phi_i} \cdot \frac{d\mathbf{x}}{d\Phi_j}\rangle, \quad \forall x \in \mathbb{R}^{3|V|}/SE(3),  (i, j)\in [0,m - 1]^2 , 
\end{equation}
where
\begin{align}
    \frac{d\mathbf{x}}{d\Phi_i} =
    \begin{cases}  
    &\frac{ \mathbf{x}_{b_i} - \mathbf{x}_{c_i}}{\| b_i - c_i \|} \times (\mathbf{x} - \mathbf{x}_{c_i}) \text{\quad if $\mathbf{x}_l \in  \mathcal{V}_{c_i} $}, \\
    &0 \text{\quad if $\mathbf{x}_l \in  \mathcal{V}_{b_i}$}, \\
    \end{cases}
\end{align}
where $(bi, ci)$ is the freely rotatable bond for torsion angle $i$, and $\mathcal{V}_{b_i}$ is the set of all nodes on the same side of the bond as $b_i$ \cite{jing2022torsional}.

This allows us to compute $dV$ as follows: 
\begin{equation}
    dV = \sqrt{det(g)} d\mathbf{x}.
\end{equation}

Thus, 
\begin{equation}
    p(\bm{x}| L, G)dx = \frac{p(\Phi| L, G)}{dV}dx = \frac{p(\Phi| L, G)}{\sqrt{det(g)}}
\end{equation}

While it is crucial to take this factor, $\sqrt{det(g)}$, into account to sample from the Boltzmann distribution, in our experiments we set $\mathrm{det}(g) = 1$ for simplicity. It is also omitted in the definition of the reward function for compatibility with our experiments. Future work will consider integrating this factor into the reward. 

\subsubsection{Forward and backward policy parametrization}
\label{supp:vonmises}

As a continuation of previous work by \citet{volokhova2024towards}, we parametrize the forward and backward policies of the GFlowNet as a mixture of von Mises distributions of $[0, 2\pi]^m$, that is:\ 
\begin{align*}
    & P^{\theta}_F(\phi_{t + 1}| \phi_t ) =   \sum_{k = 1}^{K} w^{\theta}_{k, F} \cdot \mathrm{VM}(\phi_{t + 1} | \mu^{\theta}_{k,F}(\phi_t), \kappa^{\theta}_{k, F}(\phi_t)),\\
    & P^{\theta}_B(\phi_{t}| \phi_{t + 1} ) = \sum_{k = 1}^{K} w^{\theta}_{k, B} \cdot \mathrm{VM}(\phi_{t} | \mu^{\theta}_{k,B}(\phi_{t + 1}), \kappa^{\theta}_{k,B}(\phi_{t + 1})),
\end{align*}
where
\begin{itemize}
    \item $\mathrm{VM}$ is the von Mises distribution on the torus,
    \item $K$ is the number of components of the mixture,
    \item $w^{\theta}_{k, F}$ (resp.  $w^{\theta}_{k, B}$ ) are the weights of the mixture (i.e. they are positive and sum to 1),
    \item $\mu^{\theta}_{k, F}(\phi)$ (resp. $\mu^{\theta}_{k, B}(\phi)$ ) is the location of the $k$-th component, 
    \item  $\kappa^{\theta}_{k, F}(\phi)$ (resp. $\kappa^{\theta}_{k,B}(\phi)$ ) is the concentration of the $k$-th component.
\end{itemize}

We fix the number of components, and learn the weights, locations and concentrations of each rotatable torsion angle in a molecule using VectorGNN, a new graph-neural network architecture which we describe in the following section. Using such a GNN allows to train a single GFlowNet model on multiple molecular systems, extending on previous work \cite{volokhova2024towards}, where a separate GFlowNet was trained for each molecular system, and where the policy was parametrized with an MLP.

\subsubsection{VectorGNN Architecture}\label{vectorgnn}

\paragraph{Invariant Message Passing}
The first $L$ layers of VectorGNN perform invariant message passing over a fully-connected molecular graph to compute atomic embeddings:
\begin{align}
    m^l_i &= \sum_{j = 1}^N \MLP_m(h^l_i, h^l_j, e_{ij}, \| \vec{x}_{ij} \|), \\
    h_i^{l+1} &= \MLP_h(h_i^l, m_i^l), \quad h_i^{l+1} \in \mathbb{R}^D,
\end{align}
where $\vec{x}_{ij}$ is the vector between atoms $i$ and $j$, and $e_{ij}$ denotes edge attributes. The initial atomic embeddings $h_i^0$ consist of one-hot encodings of the atom type, atomic number, one-hot encoding of the atom degree, and one-hot encoding of the atom hybridization. The edge attribute $e_{ij}$ is one-hot encoding of the bond type. Here, MLP denotes a multi-layer perceptron with GELU activations \cite{hendrycks2016gaussian}.

\paragraph{Reflection-equivariant Output}
The learned invariant embeddings are used to predict the amplitudes of interatomic forces between all atom pairs:
\begin{equation}
    f_{ij} = \MLP_f(h^L_i + h^L_j, e_{ij}, \| \vec{x}_{ij} \|) \in \mathbb{R}^{C \times 3},
\end{equation}
where $C$ is the number of force channels. These amplitudes are used to compute directional interatomic force vectors:
\[
\vec{f}_{ij} = \frac{f_{ij} \vec{x}_{ij}}{\| \vec{x}_{ij} \|^3}, \quad \vec{f}_i = \sum_{j=1}^N \vec{f}_{ij},
\]
where $\vec{f}_i$ is the net force acting on atom $i$.

For a given rotatable bond between atoms $i$ and $j$, the neighboring atoms $k \in \mathcal{N}(i)$ exert forces that generate torques about the bond:
\begin{equation}
    \vec{t}_{ij} = \sum_{k \in \mathcal{N}(i),\, k \neq j} \vec{x}_{ik} \times \vec{f}_k.
\end{equation}

Analogously, we compute $\vec{t}_{ji}$, the torque induced at atom $j$, which acts in the opposing direction to $\vec{t}_{ij}$. Since we are only interested in rotations around the bond axis, we project both torque vectors onto the bond direction and compute the net torque magnitude $s_{ij} \in \mathbb{R}^C$ as:
\begin{equation}
    t_{ij} = \vec{t}_{ij} \cdot \frac{\vec{x}_{ij}}{\| \vec{x}_{ij} \|}, \quad
    s_{ij} = t_{ij} - t_{ji}.
\end{equation}

The resulting pseudo-scalar torque magnitudes $s_{ij}$ are then passed through a reflection-equivariant MLP to produce the final output features:
\begin{gather}
    o_{ij} = \EMLP(s_{ij}) \in \mathbb{R}^{K \times 3}, \\
    \EMLP(x) = W_1 \sin(W_2 x).
\end{gather}

Because $o_{ij} = -o_{ji}$, each rotatable bond $\phi$ can be represented using either directed edge $ij$ or $ji$.

The first $K$ elements of $o_{ij}$ are passed to a softmax layer to predict the weights, the next $K$ elements correspond to the predicted locations, and the absolute value of the last $K$ elements predict the concentrations.

\subsubsection{TorsionalGFN Algorithm}

\cref{alg:torsionalGFNalgo} summarizes the TorsionalGFN training.

\begin{algorithm}[tb]
   \caption{TorsionalGFN training}
   \label{alg:torsionalGFNalgo}
\begin{algorithmic}
   \STATE {\bfseries Input:} Dataset of molecular graphs and their local structures from MD: $D = \{(G_i, L_i)\}_{  1 \leq i \leq |D| }$\\
    GFlowNet initialized with arbitrary policies $P_F^\theta$ and $P_B^\theta$\\
   Reward-prioritized ReplayBuffer \cite{vemgal2023empirical} $ \mathcal{R} = \{\mathcal{R}_i\}_{  0 \leq i \leq |D| }$ \\
    Number of training steps $N_{steps}$\\
   \FOR{$t=1$ {\bfseries to} $N_{steps}$}
   \STATE Sample a batch of graphs and local structures $B = \{(G_{k_1}, L_{k_1}), \dots, (G_{k_b}, L_{k_b})\},\quad 1 \leq b \leq |D| $
   \STATE $\mathcal{L}_D = 0$
   \FOR{$j=1$ {\bfseries to} $b$}
   \STATE  Sample a batch of trajectories of torsion angles $B_{\tau} | G_{k_j}, L_{k_j}$  using the behavior policy. $50 \%$ of the trajectories are sampled using an $\epsilon$-greedy policy \cite{jain2022biological}, with $\epsilon = 0.5$, and $50\%$ are sampled by choosing a terminal state in $\mathcal{R}_{k_j} $ and constructing a trajectory backwards with  $ P_B^{\theta}$).\\
   \STATE Compute the VarGrad loss  $\mathcal{L}_{VG}(B_{\tau}, \theta |G_{k_j}, L_{k_j} )$
   \STATE Update $\mathcal{R}_{k_j}$ with $B_\tau$, using reward-prioritization with diversity criterion \citep{vemgal2023empirical}
   \ENDFOR \\
   \STATE $ \mathcal{L}_D = \mathcal{L}_D +  \frac{1}{b}L_{VG}(B_{\tau}, \theta)$
   \STATE Backprop on $\mathcal{L}_D$ and update the parameters of $P_F^\theta$ and $P_B^\theta$ 
   \ENDFOR
\end{algorithmic}
\end{algorithm}

\subsection{Experiments}

\subsubsection{Dataset}
\begin{figure}[htbp]
\centering
\begin{subfigure}{0.22\textwidth}
  \includegraphics[width=\textwidth]{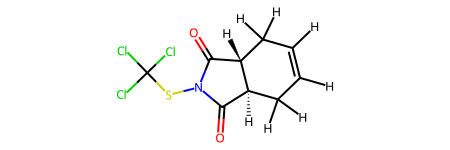}
  \caption{\tiny{C1C=CC[C@@H]2[C@@H]1C(=O)\\N(C2=O)SC(Cl)(Cl)Cl}}
  \label{fig:sub1}
\end{subfigure}
\begin{subfigure}{0.22\textwidth}
  \includegraphics[width=\textwidth]{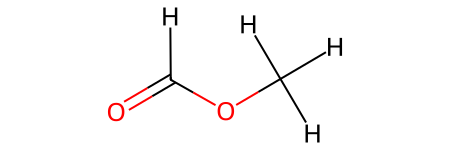}
  \caption{\tiny{COC=O}}
  \label{fig:sub2}
\end{subfigure}
\begin{subfigure}{0.22\textwidth}
  \includegraphics[width=\textwidth]{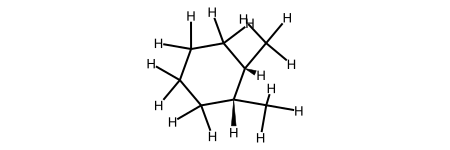}
  \caption{\tiny{C[C@@H]1CCCC[C@@H]1C}}
  \label{fig:sub3}
\end{subfigure}
\begin{subfigure}{0.22\textwidth}
  \includegraphics[width=\textwidth]{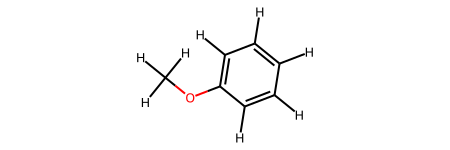}
  \caption{\tiny{COc1ccccc1}}
  \label{fig:sub4}
\end{subfigure}

\begin{subfigure}{0.22\textwidth}
  \includegraphics[width=\textwidth]{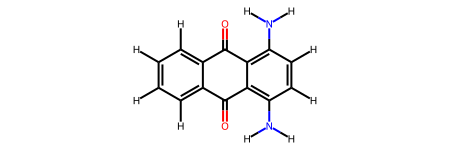}
  \caption{\tiny{c1ccc2c(c1)C(=O)c3c(ccc(c3C2=O)N)N}}
  \label{fig:sub5}
\end{subfigure}
\begin{subfigure}{0.22\textwidth}
  \includegraphics[width=\textwidth]{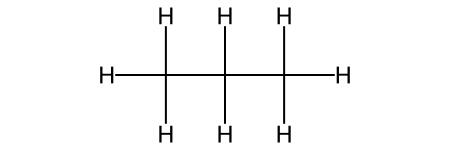}
  \caption{\tiny{CCC}}
  \label{fig:sub6}
\end{subfigure}
\begin{subfigure}{0.22\textwidth}
  \includegraphics[width=\textwidth]{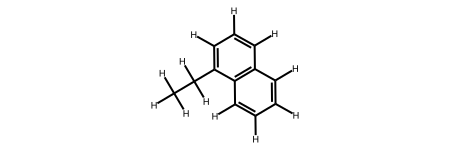}
  \caption{\tiny{CCc1cccc2c1cccc2}}
  \label{fig:sub7}
\end{subfigure}
\begin{subfigure}{0.22\textwidth}
  \includegraphics[width=\textwidth]{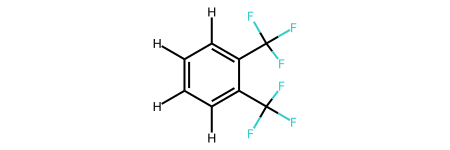}
  \caption{\tiny{c1ccc(c(c1)C(F)(F)F)C(F)(F)F}}
  \label{fig:sub8}
\end{subfigure}

\caption{Visualisation of the 8 molecular graphs in the dataset}
\label{fig:smis}
\end{figure}

Molecular graphs are visualized in Fig.\ref{fig:smis}

\subsubsection{Molecular dynamics simulation}\label{seq:app_md}

To generate conformations with Molecular Dynamics simulation, we used OpenMM \cite{eastman2017openmm} with OpenFF 2.1.1 forcefields \cite{boothroyd2023development} to compute energies in vacuum at room temperature. To get reference MD simulations we run $2\text{ns}$ simulations at a resolution of $1 \text{fs}$ and subsample to get decorrelated $1 \text{ps}$ frames. In the end, we obtain $2001$ conformations for each molecule. 
\subsubsection{Policy pre-training}
\label{supp:pretraining}
We found that training a randomly initialized VectorGNN with the GFlowNet loss required considerably more iterations than the MLP policy trained by \citet{volokhova2024towards}. To mitigate the additional training complexity of using a GNN policy, we pre-train VectorGNN on the supervised task of predicting the energy and torsion angles values of molecular conformations. To construct a dataset for this task, we use six train molecules with one corresponding local structure from MD dataset and sample 10000 values of rotatable torsion angles from the uniform distribution on a torus $[0, 2\pi]^2$ for each molecule. This gives us 60000 conformations, which we divide into 48000 for pretraining and 12000 for evaluation of the pretrained model. The model is trained to predict energy of the conformation, $\sin\phi$, and $\cos\phi$ for all rotatable torsion angles $\phi$ of a molecule.  

Then, we use the pretrained model as an initialization for training both $P_F^{\theta}$ and $P_B^{\theta}$. The the last  equivariant MLP block of VectorGNN is initialized separately without pretraining. This provides an informative initialization for the backbone VectorGNN model, which has been incentivized to extract useful representations for predicting energy and shape of the conformations -- a desirable inductive bias to facilitate Torsional-GFN training.

\subsubsection{Comparison of the local structures distribution with MD simulations and RDKit}\label{md_rdkit_localstructures}

\begin{figure}[H]
\centering     
\begin{subfigure}{0.48\textwidth}
  \includegraphics[width=\textwidth]{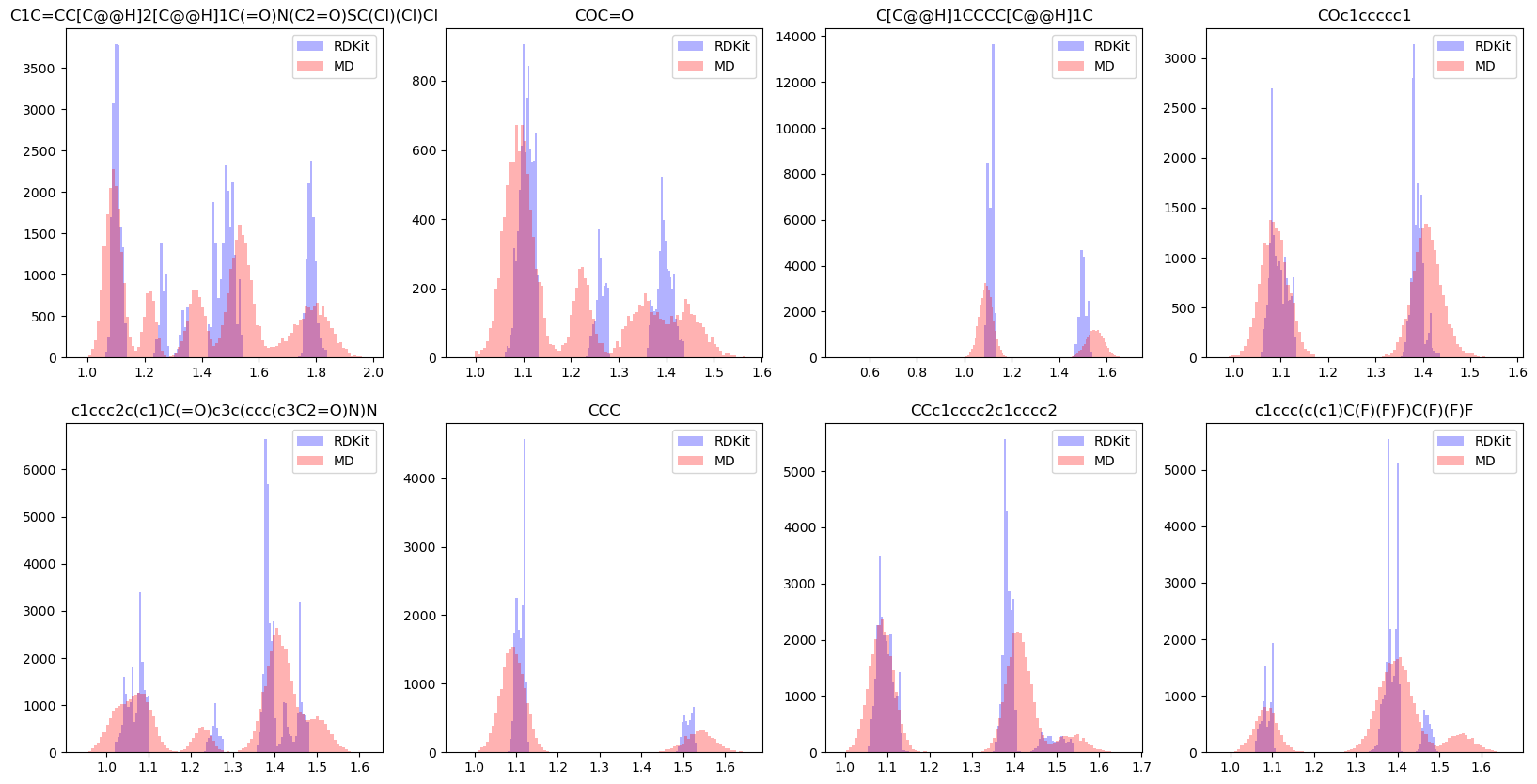}
  \caption{Bond lengths}
  \label{fig:bonglenthshist}
\end{subfigure}
\begin{subfigure}{0.48\textwidth}
  \includegraphics[width=\textwidth]{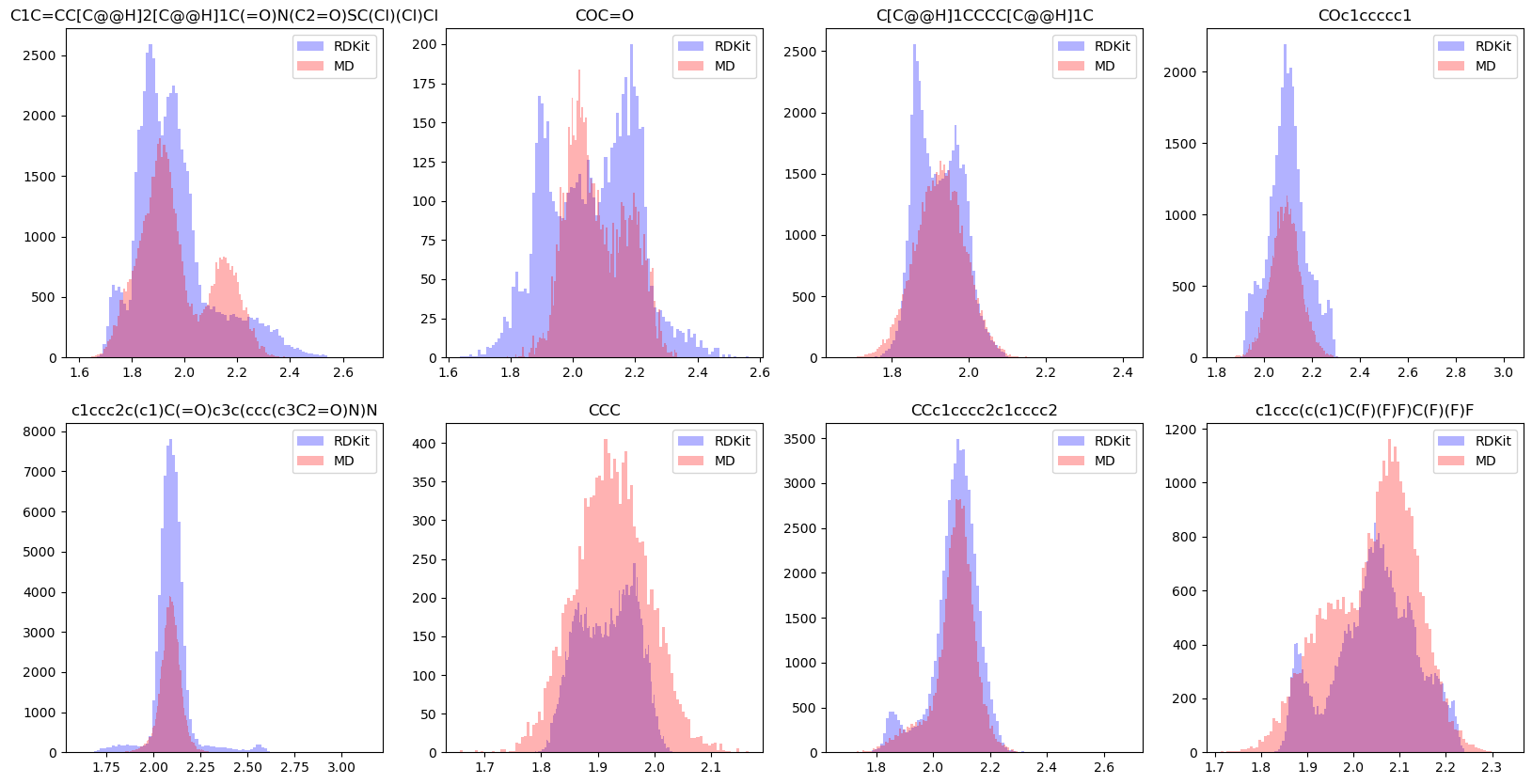}
  \caption{Bond angles}
  \label{fig:bondangles hist}
\end{subfigure}
\caption{Histograms of local structures of MD data (pink) compared to local structures generated by RDKit (purple) for the 8 molecules in our dataset. Left is bond lengths, right is bond angles. Note that RDKit significantly underestimated the vibrations of the bond lengths, resulting in peakier modes for all molecules.}
\label{fig:localstruct_hist}
\end{figure}

The Figure \ref{fig:localstruct_hist} visualizes distribution of the values of the bond lengths and angles in the conformations sampled with MD and with RDKit. Noticeably, the distributions for MD conformations have smooth peaks with significant variance. This confirms that local structures are not fixed to specific values in the ground truth Boltzmann distribution, but follow a smooth distribution due to temperature vibrations of atoms in a molecule.

\subsubsection{Correlation between sampling probability and reward}\label{seq:app_corr}

To estimate the sampling log-probabilities of the trained Torsional-GFN model, we follow the algorithm described in \cite{volokhova2024towards}. We use $N=10$ trajectories for each estimation.

\begin{equation}\label{eq:corr_coeff}
\rho_{\log p_\top^{\theta}, log R} = \frac{\mathrm{cov}(\log p_\top^{\theta}(\Phi_i | G, L), \log R(\Phi_i | G, L))}{\sigma_{\log p_\top^{\theta}(\Phi_i | G, L)}\sigma_{\log R(\Phi_i | G, L)}}.
\end{equation}

\begin{figure}[ht]
\centering
\begin{subfigure}{\textwidth}
    \includegraphics[width=\textwidth]{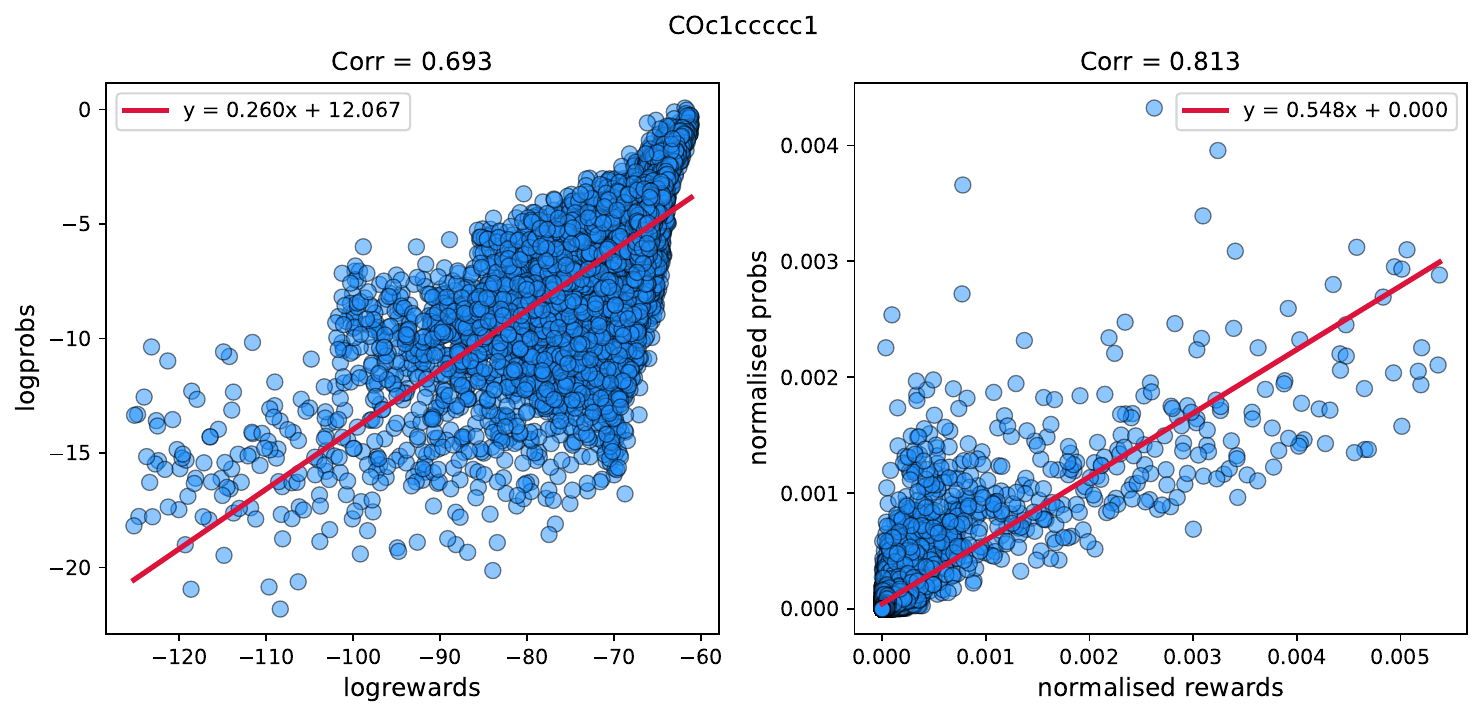}
\end{subfigure}
\hfill
\begin{subfigure}{\textwidth}
    \includegraphics[width=\textwidth]{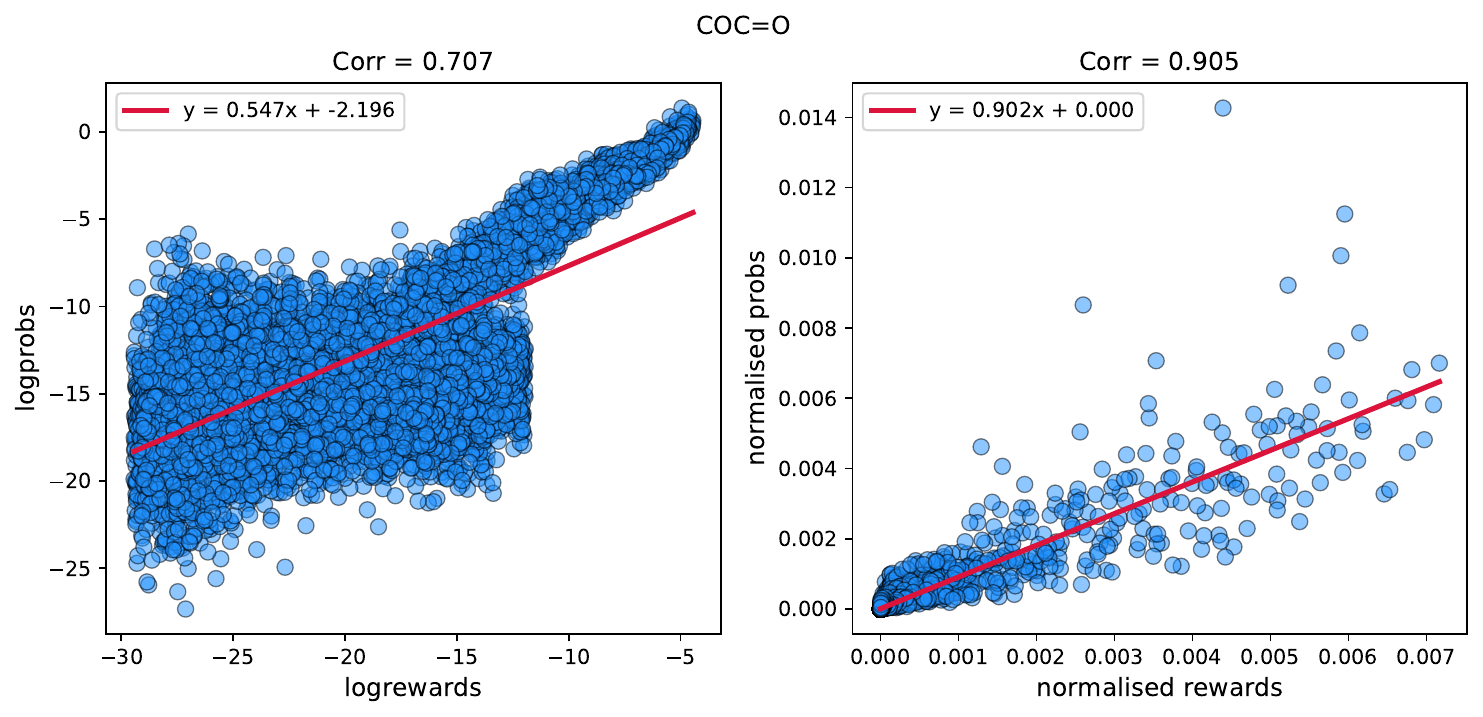}
\end{subfigure}
\caption{Examples of the scattered plots of rewards and sampling probabilities in the log scale and without it. The values are computed on a grid of 10,000 torsion angles values. As one can see, the correlation coefficient of the direct values is higher than for the logarithmic ones due to the low-reward values collapsing in the corner close to zero. This pattern shows that correlations of the direct values can underestimate the discrepancy of the sampling probability and reward in the low reward regions.}
\label{fig:corrs}
\end{figure}

To measure correspondence between sampling probabilities and rewards we use correlation coefficient of their logarithms as defined in Eq. \ref{eq:corr_coeff}. 
The choice of the logarithms instead of the direct values is motivated by the observation that logarithmic scale allows to mitigate the over-focus on the high-reward samples, while correlation of the direct values is often biased by the high-reward samples (see for example \cref{fig:corrs}), especially for the molecules with sharp changes in the energy landscape. 

\subsubsection{2D visualisations for train molecules}

\begin{figure}[htbp]
\centering
\begin{subfigure}{0.48\textwidth}
  \includegraphics[width=\textwidth]{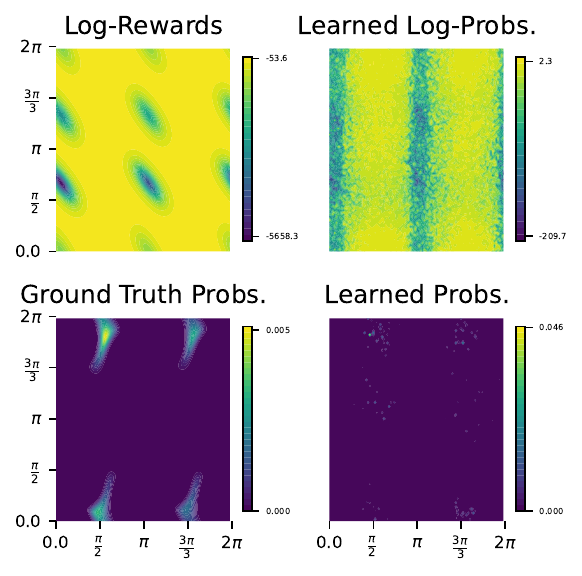}
  \caption{\tiny{Train}}
\end{subfigure}
\begin{subfigure}{0.48\textwidth}
  \includegraphics[width=\textwidth]{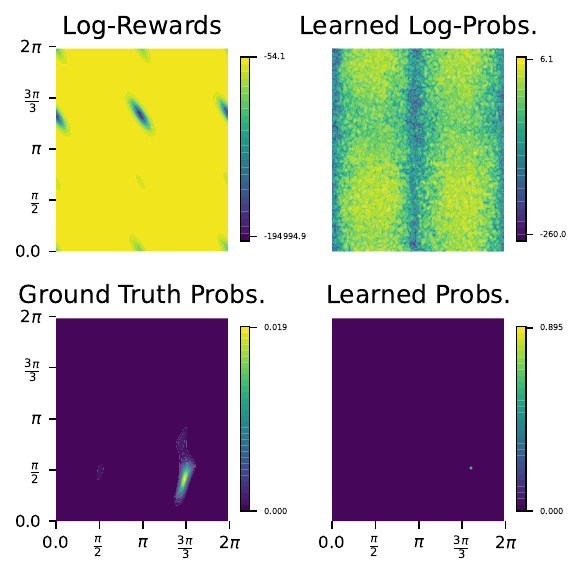}
  \caption{\tiny{Unseen}}
\end{subfigure}
\caption{Visualization of the log-rewards and ground truth probabilities alongside the learned sampling distributions for molecule C1C=CC[C@@H]2[C@@H]1C(=O)N(C2=O)SC(Cl)(Cl)Cl in the train dataset, for train and unseen local structures.}
\label{fig:2dviztrain_mol1}
\end{figure}

\begin{figure}[htbp]
\centering
\begin{subfigure}{0.48\textwidth}
  \includegraphics[width=\textwidth]{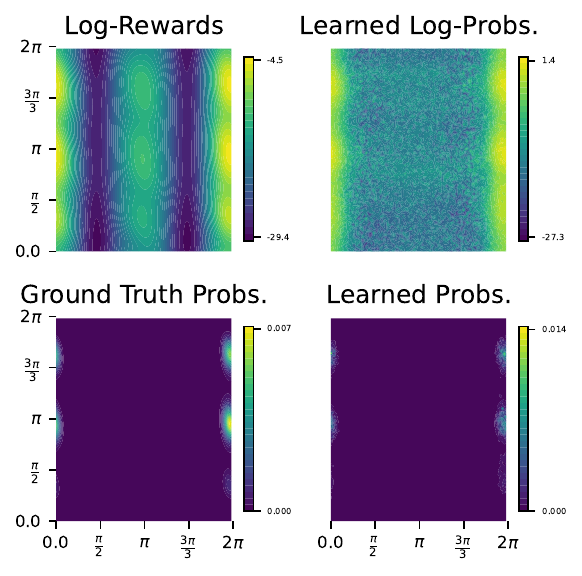}
  \caption{\tiny{Train}}
\end{subfigure}
\begin{subfigure}{0.48\textwidth}
  \includegraphics[width=\textwidth]{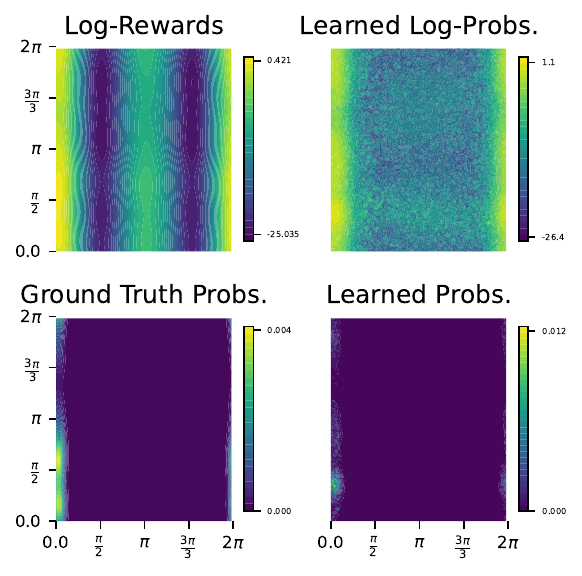}
  \caption{\tiny{Unseen}}
\end{subfigure}
\caption{Visualization of the log-rewards and ground truth probabilities alongside the learned sampling distributions for molecule COC=O in the train dataset, for train and unseen local structures.}
\label{fig:2dviztrain_mol2}
\end{figure}

\begin{figure}[htbp]
\centering
\begin{subfigure}{0.48\textwidth}
  \includegraphics[width=\textwidth]{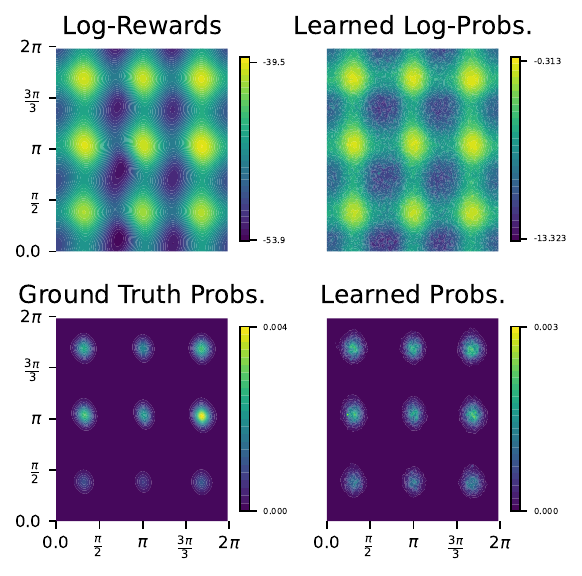}
  \caption{\tiny{Train}}
\end{subfigure}
\begin{subfigure}{0.48\textwidth}
  \includegraphics[width=\textwidth]{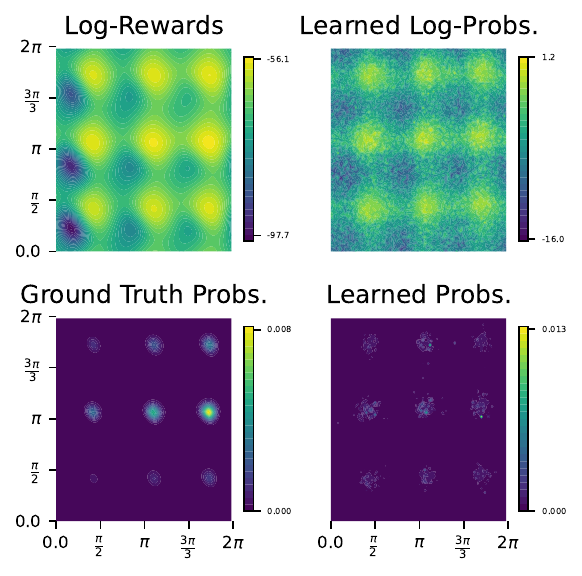}
  \caption{\tiny{Unseen}}
\end{subfigure}
\caption{Visualization of the log-rewards and ground truth probabilities alongside the learned sampling distributions for molecule C[C@@H]1CCCC[C@@H]1C in the train dataset, for train and unseen local structures.}
\label{fig:2dviztrain_mol3}
\end{figure}

\begin{figure}[htbp]
\centering
\begin{subfigure}{0.48\textwidth}
  \includegraphics[width=\textwidth]{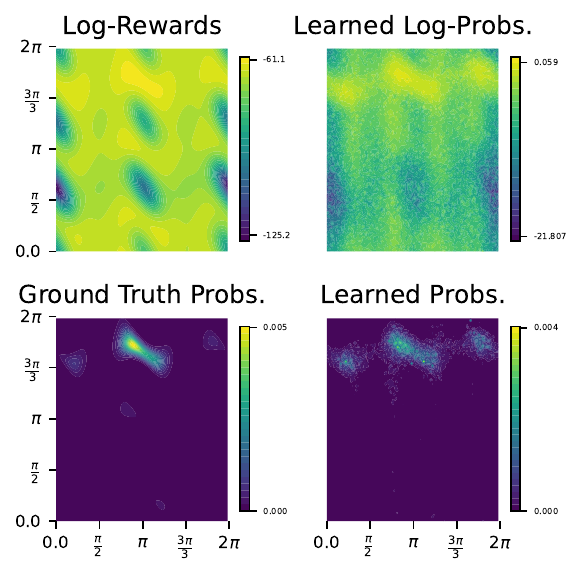}
  \caption{\tiny{Train}}
\end{subfigure}
\begin{subfigure}{0.48\textwidth}
  \includegraphics[width=\textwidth]{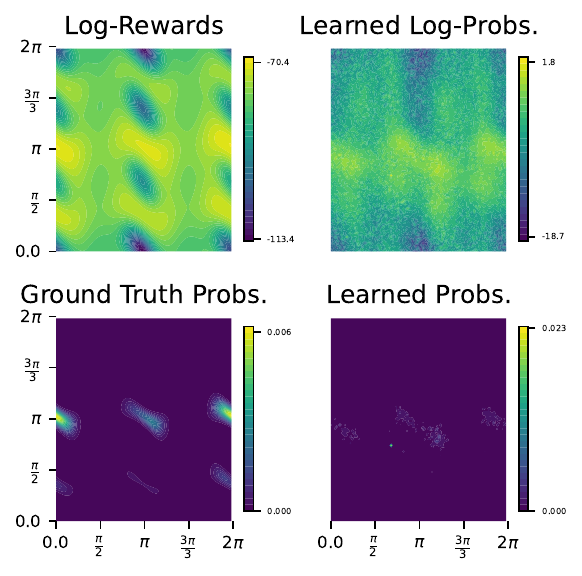}
  \caption{\tiny{Unseen}}
\end{subfigure}
\caption{Visualization of the log-rewards and ground truth probabilities alongside the learned sampling distributions for molecule c1ccc2c(c1)C(=O)c3c(ccc(c3C2=O)N)N in the train dataset, for train and unseen local structures.}
\label{fig:2dviztrain_mol4}
\end{figure}

\begin{figure}[htbp]
\centering
\begin{subfigure}{0.48\textwidth}
  \includegraphics[width=\textwidth]{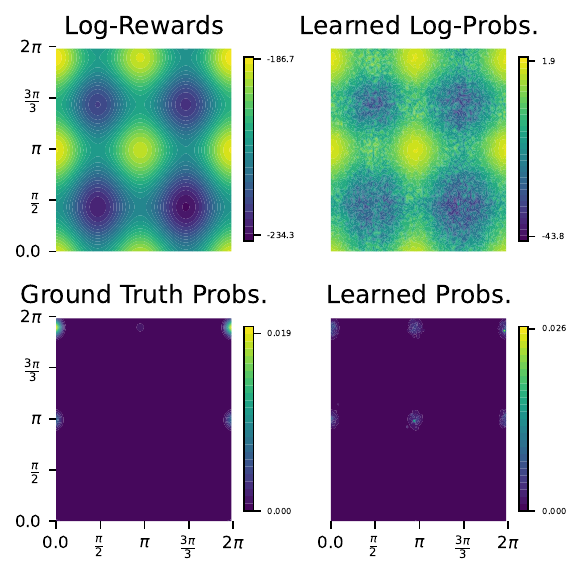}
  \caption{\tiny{Train}}
\end{subfigure}
\begin{subfigure}{0.48\textwidth}
  \includegraphics[width=\textwidth]{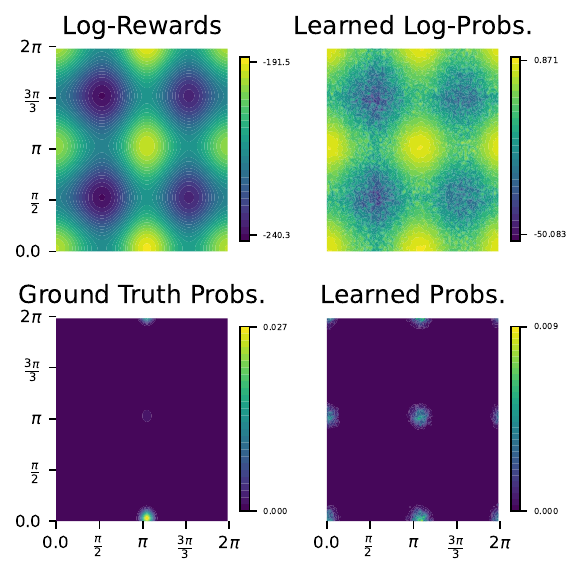}
  \caption{\tiny{Unseen}}
\end{subfigure}
\caption{Visualization of the log-rewards and ground truth probabilities alongside the learned sampling distributions for molecule CCC in the train dataset, for train and unseen local structures.}
\label{fig:2dviztrain_mol5}
\end{figure}

This section shows additional visualisations of the 2D log-rewards alongside the learned distributions in \cref{fig:2dviztrain_mol1,fig:2dviztrain_mol2,fig:2dviztrain_mol3,fig:2dviztrain_mol4,fig:2dviztrain_mol5}.

\subsubsection{2D visualizations for test molecules}

\begin{figure}[htbp]
    \centering
    \begin{subfigure}[b]{0.45\textwidth}
        \centering
        \includegraphics[width=\textwidth]{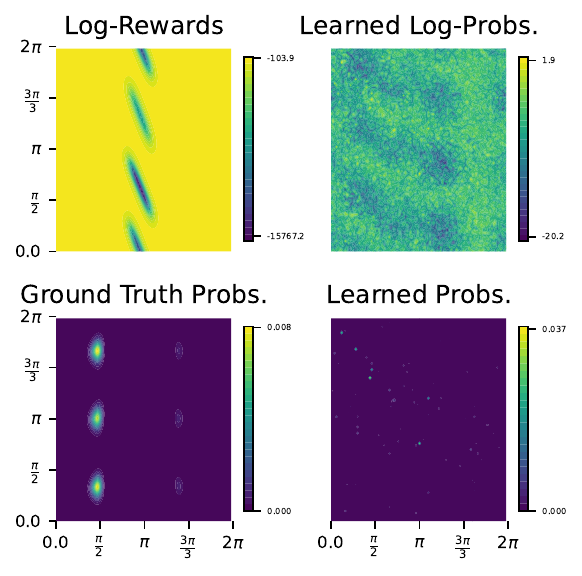}
        \caption{CCc1cccc2c1cccc2}
        \label{fig:sub1}
    \end{subfigure}
    \hfill
    \begin{subfigure}[b]{0.45\textwidth}
        \centering
        \includegraphics[width=\textwidth]{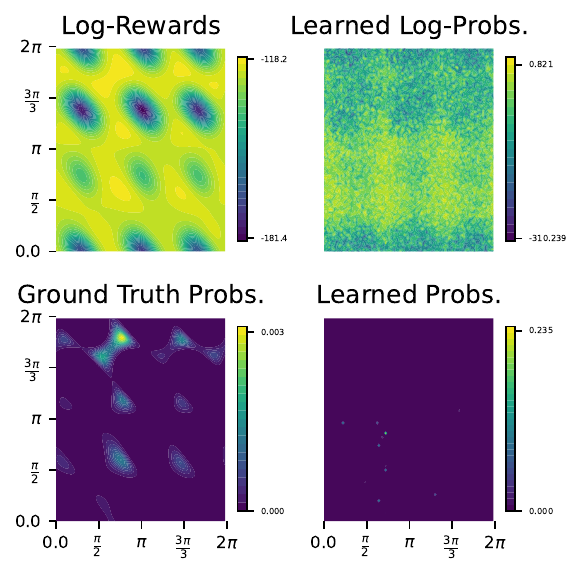}
        \caption{c1ccc(c(c1)C(F)(F)F)C(F)(F)F}
        \label{fig:sub2}
    \end{subfigure}
    \caption{Visualization of the log-rewards and ground truth probabilities alongside the learned sampling distributions for 2 molecules in the test dataset.}
    \label{fig:2dviztest}
\end{figure}

\end{document}